\documentclass{IOS-Book-Article}
\usepackage[numbers]{natbib}
\usepackage{mathptmx}
\usepackage{soul}\setuldepth{article}
\usepackage{enumitem}
\def\hb{\hbox to 11.5 cm{}}

\usepackage{times}
\usepackage{latexsym}
\usepackage{url}
\usepackage{covington}
\usepackage{booktabs}
\usepackage{graphicx}
\usepackage{amsmath} 
\usepackage{multirow} 
\usepackage{subfigure}
\usepackage{placeins}

\usepackage[T1]{fontenc}

\usepackage[utf8]{inputenc}

\usepackage{microtype}

\usepackage{inconsolata}
\def\hb{\hbox to 11.5 cm{}}

\begin{document}

\pagestyle{headings}
\def\thepage{}
\begin{frontmatter}              

\title{Towards Automating Text Annotation: A Case Study on Semantic Proximity Annotation using GPT-4\\}

\markboth{}{April 2024\hb}
\markboth{}{April 2024\hb}

\author{\fnms{Sachin} \snm{Yadav}%
},
\author{\fnms{Tejaswi} \snm{Choppa}}
and
\author{\fnms{Dominik} \snm{Schlechtweg}}
\runningauthor{sachin et al.}
\address{Institute for Natural Language Processing, University of Stuttgart}
\texttt{{st185750@stud.uni-stuttgart.de, st180670@stud.uni-stuttgart.de, {dominik.schlechtweg@ims.uni-stuttgart.de}}}

%

\begin{abstract}
This paper explores using GPT-3.5 and GPT-4 to automate the data annotation process with automatic prompting techniques. The main aim of this paper is to reuse human annotation guidelines along with some annotated data to design automatic prompts for LLMs, focusing on the semantic proximity annotation task. Automatic prompts are compared to customized prompts. We further implement the prompting strategies into an open-source text annotation tool, enabling easy online use via the OpenAI API. Our study reveals the crucial role of accurate prompt design and suggests that prompting GPT-4 with human-like instructions is not straightforwardly possible for the semantic proximity task. We show that small modifications to the human guidelines already improve the performance, suggesting possible ways for future research. 

\end{abstract}
\end{frontmatter}
\markboth{April 2024\hb}{April 2024\hb}

\section{Introduction}
Data annotation, the process of labeling data plays an important role for training machine learning models. High-quality data annotations ensure the model grasps the relationship between input data and desired output \cite{Ding2022IsGA}. However, data annotation is complex, expensive, and time-consuming. Data can be ambiguous, subjective, and exist in diverse formats, often requiring domain expertise for accurate annotation \cite{Tan2024LargeLM}. Studies have shown that advanced Large Language Models (LLMs) such as GPT-4 \cite{OpenAI2023GPT4TR}, Gemini \cite{Anil2023GeminiAF} and Llama-2 \cite{Touvron2023Llama2O} offer a promising opportunity to revolutionize data annotation. \cite{Gilardi2023ChatGPTOC} discovered that LLMs are approximately thirty times more cost-effective than human annotators, with a cost per annotation of less than \textdollar{0.003}. Additionally, they exhibit a remarkable speed advantage over human annotators. They further offer human-like performance in various downstream tasks. To get the most out of the system, users provide instructions in natural language (\textbf{prompts}), and the system responds accordingly \cite{Brown2008,Li2022LearningTT}. A prompt serves as a set of instructions guiding LLMs, refining their capabilities to suit the specific task \cite{Liu2021PretrainPA,White2023APP}. Prompting has shown promising results in low-resource settings and can effectively extract knowledge from pre-trained language models \cite{Goswami2023SwitchPromptLD}. Recent research suggests that prompting techniques can be effectively used with LLMs to automate or assist with annotation tasks \cite{Huang2023IsCB,Ding2022IsGA}. However, writing an effective prompt is crucial as quality prompts generate quality output, as discussed in \cite{White2023APP}. Manual prompt creation is further time-consuming, sophistic, and can lead to inconsistent, volatile or counter-intuitive results \cite{Shin2020ElicitingKF,sclar2023quantifying,li2023large}. We aim to utilize existing resources, such as human annotation guidelines and small sets of well-annotated data ("gold data"), to automatically prompt LLMs in order to make the annotation process as easy as possible. An essential part of our study is the implementation of the automatic prompting strategies into an open-source annotation tool exploiting the OpenAI API to make it possible to do automatic prompting with few clicks in an online user interface.

\section {Related Work}
\subsection{Text Annotation}

Human language is complex, requiring machines to understand linguistic elements. Text annotation bridges this gap by labeling text data, providing additional information like highlighting specific elements, assigning categories, or adding comments. This labeled data is essential for training accurate machine learning models that can analyze and understand human language effectively. Without text annotation, machines would struggle with the nuances of language, leading to inaccurate or meaningless results. Several general-purpose text annotation tools are available, offering flexibility for defining various tasks. These tools include CATMA \cite{evelyn_gius_2022_6419805}, INCEpTION \cite{klie-etal-2018-inception}, MTurk\footnote{\url{https://www.mturk.com/}}, PhiTag\footnote{\url{https://phitag.ims.uni-stuttgart.de/}}, Toloka\footnote{\url{https://toloka.ai/docs/}} or POTATO \cite{pei-etal-2022-potato}. 

\subsection{Large Language Models} 
LLMs have demonstrated impressive capabilities in understanding and processing natural language tasks. These models are pre-trained on massive amounts of text data that enables them to learn the patterns, structures, and relationships within language. Also, LLMs have a large number of parameters that help it to learn the complex linguistic patterns to generate responses with remarkable speed and accuracy \cite{Brown2020GPT3}. 

\begin{figure*}[h]
    \centering
    \includegraphics[width=0.9\textwidth]{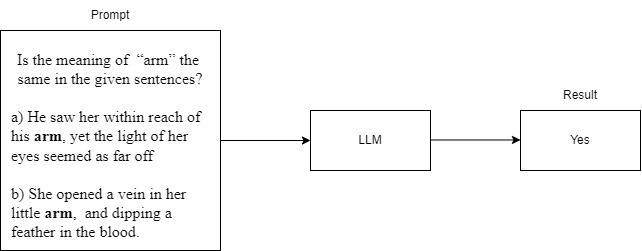}
    \caption{Prompting LLM for use pair meaning task.}
    \label{fig:ex-prompt}
\end{figure*}

\paragraph{Prompting LLMs} 
Prompting refers to the process of interacting with LLMs through textual content. A prompt typically serves as a crucial medium for engaging with LLMs and guiding them to perform specific tasks or generate desired outputs \cite{Liu2023PromptingFF}. A very simple example of prompting can be seen in Figure \ref{fig:ex-prompt}.

\paragraph{LLMs as Annotators}
LLMs demonstrate potential as data annotators for various textual tasks \cite{Karjus2023MachineassistedMM,Ding2022IsGA,Huang2023IsCB}. In the evaluations conducted by \cite{Guo2023HowCI}, key differences between ChatGPT and human experts emerged. ChatGPT's responses exhibit a strict focus on given questions, objectivity, formality, and minimal emotional expression, while human responses tend to diverge, including subjective expressions, exhibit colloquialism, and convey emotions through punctuation and language features. These distinctions highlight the unique strengths and limitations of ChatGPT in comparison to human expertise.

\cite{Huang2023IsCB} evaluated ChatGPT's proficiency in detecting and explaining implicit hate speech using the Latent Hatred data-set \cite{elsherief-etal-2021-latent}. The study revealed that ChatGPT's ability to detect implicit hate speech was on par with human capabilities, and the quality of its generated explanations closely mirrored human-written ones. This emphasizes LLMs effectiveness as a valuable tool for data annotation. This finding is particularly interesting for lexical semantic tasks like Lexical Semantic Change Detection \cite{schlechtweg-etal-2020-semeval} or Use Pair Semantic Proximity Annotation \cite{Erk13}.

\cite{Karjus2023MachineassistedMM} and \cite{periti2024chatgptvbertdawn} demonstrated that GPT-4 and GPT-3.5 can handle challenging deep semantic tasks like detecting semantic shifts in words over time (Lexical Semantic Change Detection) and meaning distinction between word uses (Use Pair Semantic Proximity Annotation). Semantic analysis, crucial for NLP, delves into the meaning and context of text, analyzing word relationships to understand the overall message. This includes further tasks like sentiment analysis and information extraction, aiming to achieve accurate text interpretation. 

We choose semantic proximity annotation as a target task in this study because it is simple, gold data and high-quality annotation guidelines are available in the annotation interface and it is very relevant for semantic applications of NLP. 

\section{PhiTag}
\subsection{Idea}
\begin{figure*}[t]
    \centering
    \includegraphics[width=0.9\textwidth]{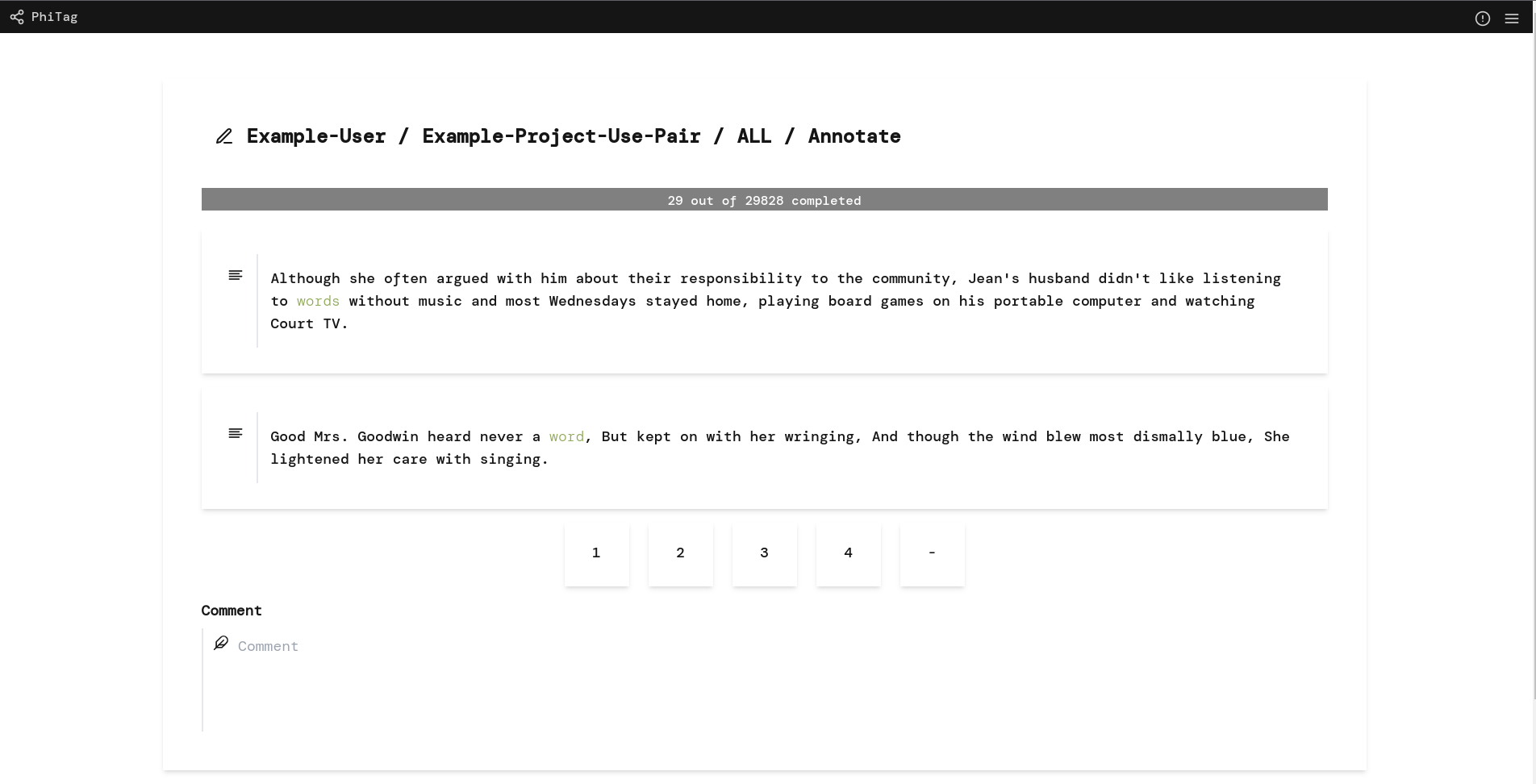}
    \caption{PhiTag text pair annotation page.}
    \label{fig:phitag}
\end{figure*}

PhiTag, an open-source text annotation platform, addresses the critical need for high-quality training data in machine learning, particularly for NLP tasks.\footnote{\url{https://github.com/Garrafao/phitag}} Recognizing the importance of custom data labeling and annotation quality, PhiTag offers a user-friendly environment designed for flexibility and efficient data management. The platform provides support for various general NLP tasks such as text pair, text ranking or text labeling annotation. The platform provides annotation guidelines, data for annotator checks (tutorial), an annotation interface and agreement calculation to ensure that users understand the specific requirements and can become proficient annotators.\footnote{\url{https://phitag.ims.uni-stuttgart.de/guide/explained-annotation-task-urel}} We utilize PhiTag's text pair annotation task type for our study. A screenshot showcasing some text pair instances can be found in Figure~\ref{fig:phitag}.

\subsection{Computational Annotator}
\begin{figure*}[t]
    \centering
    \includegraphics[width=0.9\textwidth]{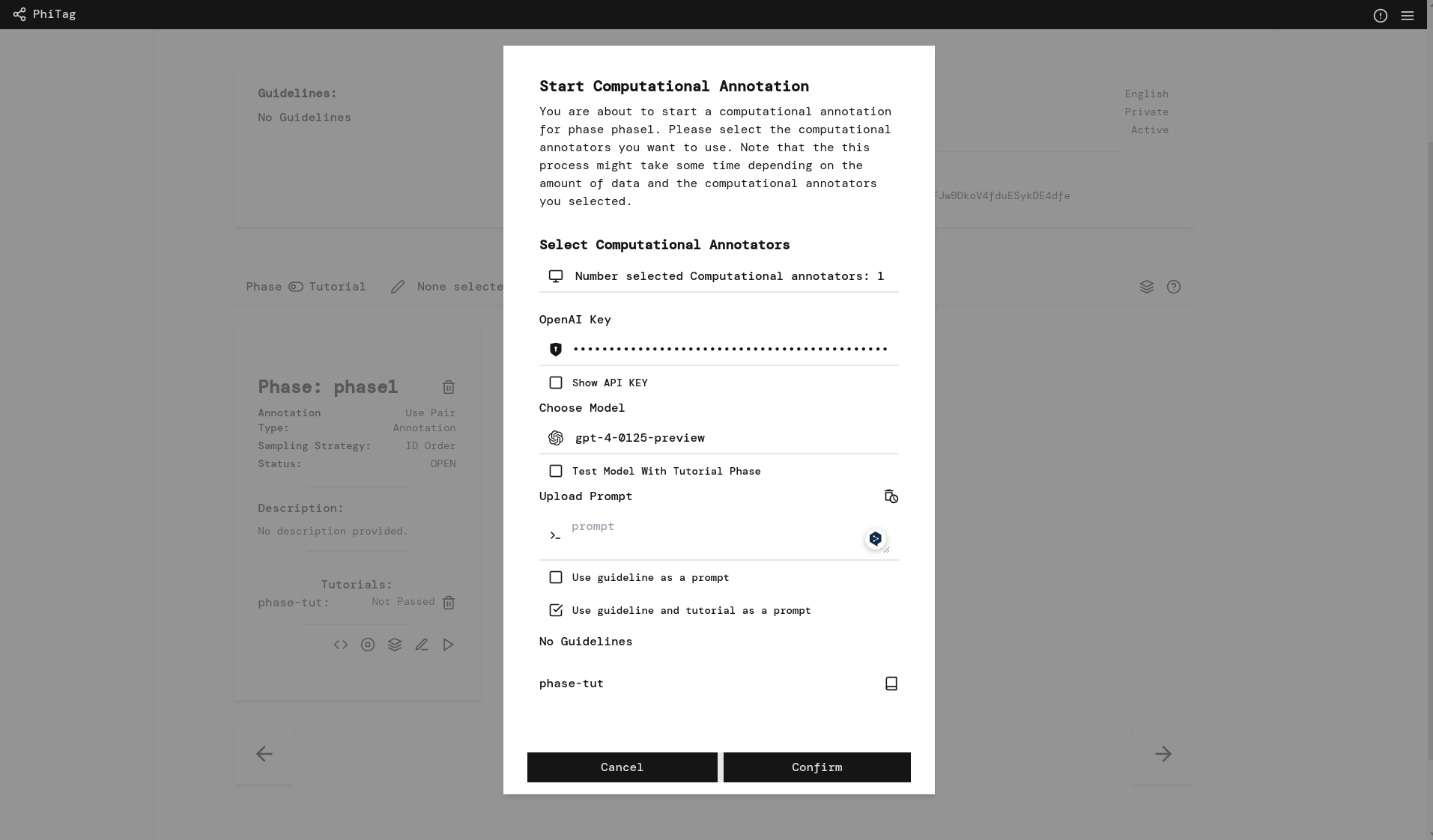}
    \caption{PhiTag computational annotator interface.}
    \label{fig:phitag-computer}
\end{figure*}

We implement a computational annotator in PhiTag which automatically annotates text data using the power of LLMs. We use the API provided by OpenAI.\footnote{\url{https://openai.com/}} This allows us to use different types of models (\texttt{gpt-4-0125-preview} used in this study) available in OpenAI and to define different types of prompts including (i) the upload of a custom prompt instruction (ii) automatic prompting features combining the annotation guidelines and tutorial data already available in PhiTag. This allows us to automate the annotation process as far as possible by reusing existing instructions and data. Find a screenshot of the computational annotator interface in PhiTag in Figure \ref{fig:phitag-computer}.

\section{Task Description}
Given the use pairs (often two sentences) of a word, our task of choice (Use Pair Semantic Proximity Annotation) involves evaluating the semantic relatedness between the two uses. Annotators judge pairs of sentences containing a highlighted target word, rating the meaning relatedness on a 4-point scale: 1 (unrelated), 2 (distantly related), 3 (closely related), and 4 (identical) \cite[][]{Schlechtwegetal18}. They first determine the most likely meaning for the target word in each sentence independently, then assess the connection between those meanings. 
 
As examples, consider (\ref{ex: 1}) (relatedness of "bank") and (\ref{ex: 2}) (relatedness of "eat").
   \begin{examples}
    \item \textbf{Sentence 1:} His parents had left a lot of money in the \textbf{bank} and now it was all Measle's, but a judge had said that Measle was too young to get it. \label{ex: 1}
    
     \textbf{Sentence 2:} Sherrell, is is said, was sitting on the \textbf{bank} of the river close by, and as soon as the men had disappeared from sight he jumped on board the schooner.

      \textbf{Target word:} bank
      
      \textbf{Judgement:} 1 (Unrelated)
    \end{examples}
    \begin{examples}
    \item \textbf{Sentence 1:} Speaking of bread and butter reminds me that we’d better \textbf{eat} ours before
    the coffee gets cold.\label{ex: 2}
    
     \textbf{Sentence 2:} When the meal was over and they had finished their tea after they \textbf{ate}, wang the Second took the trusty man to his elder brother’s gate.

      \textbf{Target word:} eat
      
      \textbf{Judgement:} 4 (Closely related)
    \end{examples}
    
\section{Data}
\label{sec:data}
Our study is based on the publicly available DWUG EN "Use Pair" dataset (V2.0.0) \cite{Schlechtweg2021dwug}.\footnote{\url{https://github.com/ChangeIsKey/annotation_standardization/tree/main/use_pair/urel/english/data}} The dataset contains 46,000 human judgments of use pairs according to the semantic relatedness scale described above.

\paragraph{Filtering, Cleaning and Splitting}
To ensure high-quality data, we employ a strict filtering condition keeping only instances without "Cannot decide" judgment and annotated by at least 2 human annotators. We include only data points where all annotators agreed. This filtering process resulted in a final dataset of approximately 930 instances. We employ a random sampling technique to split the gold data into three distinct sets: training, testing, and development. The specific distribution of instances can be found in Table \ref{tab:data}, and the distribution of labels of each data set can be found in Figure \ref{fig:label-distribution-data-gold}.

\begin{table}[t]
    \centering
    \begin{tabular}{c|c|c} 
        \toprule
        \textbf{Dev} & \textbf{Train} & \textbf{Test} \\ 
        \midrule
        46 & 140 & 744 \\
        \bottomrule
    \end{tabular}
\vspace{0.2cm}
\caption{Test data split statistics}
    \label{tab:data}
\end{table}

\section{Experiments}
We now describe our experiments on the data described in Section \ref{sec:data}. We start out by fixing model parameters at their default value and optimizing the custom prompt on the development data. We then test the influence of model parameters using the optimized custom prompt. We continue by using these optimal model parameters to optimize prompts for the fine-tuned model and automatic prompting on the development data. We then test the optimal configurations for each strategy on the test data and report performance. We provide the data, code and prompts in a public repository.\footnote{\url{https://github.com/sachin1022/phitag_gpt_datasets}}

\paragraph{Performance Metrics} 
Performance between the gold data and the computationally annotated data in our experiment is measured by two metrics: ordinal Krippendorff's $\alpha$ \cite{krippendorff2018content} and percentage agreement. Percentage agreement measures the normalized share of agreements among the total number of annotations. However, it does not account for chance agreement and graded deviation from gold. In contrast, Krippendorff's $\alpha$ offers a more robust assessment. This versatile statistic considers both observed agreement and agreement expected by chance, providing a more reliable picture of inter-annotator agreement. Krippendorff's alpha can handle various data types (nominal, ordinal, interval, and ratio), and a score of 1 indicates perfect agreement \cite{10.1162/coli.07-034-R2}.

\subsection{Customized Prompt}
\label{customized-prompt}

\paragraph{Prompt} The customized prompt consists of an instruction message to the system, a custom message consisting of the two uses to be judged along with the target lemma and the message to obtain the judgment. Find an example in Appendix \ref{app:prompt}. We explore prompting strategies to optimize performance of the model. Initial attempts were done by using basic prompts such as "Your task is to rate the degree of semantic relatedness between two uses of a target word in the given sentences" inspired by the human task guidelines. See Appendix \ref{ex:prompt-1} for complete prompt. Inspired by the work of \cite{Karjus2023MachineassistedMM}, we also adopted and refined their prompt structure. The instruction was modified by adding the following sentence towards the end of prompt: "Your response should align with a human's succinct judgment." This modification aimed to guide the model towards human-like evaluations of semantic relatedness. To align the generated annotation with the semantic relatedness scale, we further polish the prompt by adding the sentence "Please provide a judgment as a single integer. For example, if your judgment is Identical, then provide 4. If your judgment is Unrelated, provide 1." See Appendix \ref{app:prompt} Prompt 2 for the final prompt.

\begin{figure}[t]
\centering
\subfigure[Train Data]{
\includegraphics[width=0.3\textwidth]{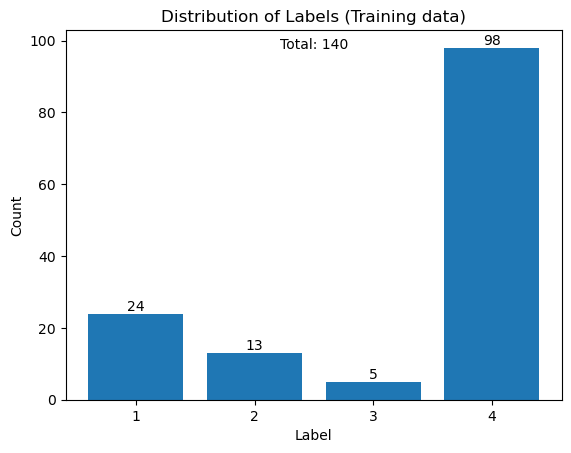}}
\subfigure[Test Data]{
\includegraphics[width=0.3\textwidth]{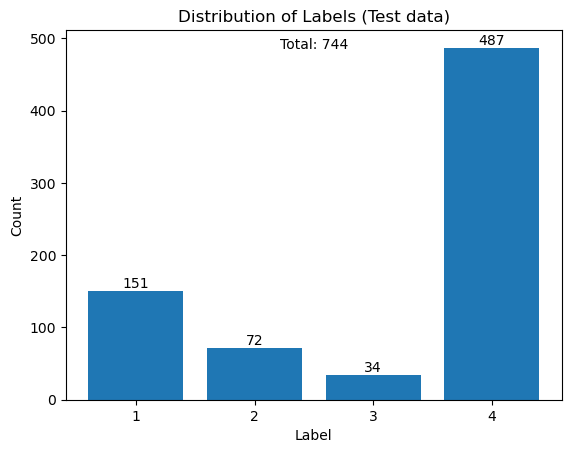}}
\subfigure[Dev Data]{
\includegraphics[width=0.3\textwidth]{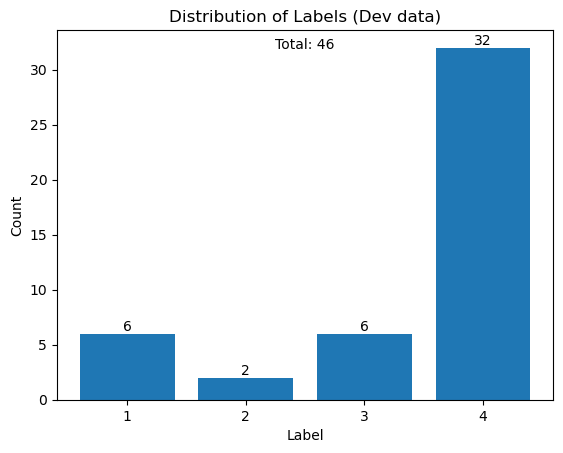}}
\caption{Distribution of labels in gold data.}
\label{fig:label-distribution-data-gold}
\end{figure}

\paragraph{Model Development} After fixing the prompt, we experiment with \texttt{gpt-4-0125-preview} from the GPT-4 class, the latest version specifically designed to address "laziness" issues. This typically arises when the model fails to complete tasks adequately, either by providing incomplete or insufficient responses.\footnote{\url{https://platform.openai.com/docs/models/continuous-model-upgrades}} The model's performance is affected by several parameters such as \texttt{temperature}, \texttt{top-p}, \texttt{stop-condition}, \texttt{max-token}, etc. 
We examine the impact of the \texttt{temperature} and \texttt{top-p} parameters on the performance of GPT-4 model outputs using the development data. These parameters influence the diversity and randomness of the model's output which in turn indicate the quality of the generated responses \cite{peng2023making, wang-etal-2021-transprompt}.

Higher \texttt{top-p} values allow for more diverse and creative output and vice-versa. Likewise, higher \texttt{temperatures} introduce more randomness and exploration, resulting in potentially more diverse but less coherent output. We explore the impact of \texttt{temperature} values ranging from min 0.1 to max 1.0 at steps of 0.1, and \texttt{top-p} values (0.1 to 1.0) in steps of 0.1. The primary objective is to determine the optimal model configuration by adjusting these parameters. We found that a \texttt{temperature} setting of 0.9 and a \texttt{top-p} value of 0.9 yielded the best performance for our study with Prompt 2 (see Appendix \ref{app:prompt}).

Upon various trials conducted on the above-described model configuration and selected optimized prompt, we find the optimal performance of the model at mean $\alpha$ of 0.74 and percentage agreement of 0.80 achieved across 5 different trials as shown in Table \ref{tab:dev-data-results}.

\paragraph{Model Testing}

After finalizing model development, we evaluate performances on a test dataset (gold data) of 744 instances as displayed in Table \ref{tab:data} using both Krippendorff's $\alpha$ and percentage agreement. In the test data, the instances labelled with 4 are most frequent and instances labelled with 3 are least frequent (see Table~\ref{fig:label-distribution-data-gold} for a detailed distribution). We achieve consistent results across multiple trials. Throughout five trials, we get consistent results with mean $\alpha$ of 0.54 and 0.72 for percentage agreement, as shown in Table~\ref{tab:dev-data-results}.

\begin{table}[t]
    \centering
    \begin{tabular}{c|cc|cc} 
        \toprule
         \multirow{2}*{\textbf{Trial}}   & \multicolumn{2}{c|}{\textbf{dev}} & \multicolumn{2}{c}{\textbf{test}}\\     & $\alpha$ & $\%$& $\alpha$ & $\%$ \\ 
        \midrule
         1 & 0.74 & 0.80 & 0.56 & 0.73\\
         2 & 0.75  & 0.81 & 0.54 & 0.71\\
         3 & 0.75  &0.80 & 0.54 &0.72\\
         4 & 0.74  &0.80& 0.54 &0.72\\
         5 & 0.74  &0.80& 0.53 &0.73\\
        \midrule
        \textbf{Mean} & 0.74 & 0.80 & 0.54 & 0.72 \\
        \bottomrule
    \end{tabular}
    \vspace{0.2cm}
    \caption{Custom prompting results. $\alpha$: Krippendorff's $\alpha$, $\%$: Percentage agreement.}
    \label{tab:dev-data-results}
\end{table}

\subsection{Fine-Tuning}
Fine-tuning a large language model involves training the model further on a smaller, targeted dataset that is relevant to the desired task or subject matter. This can also be done through prompting. In the study conducted by \cite{Sun2023DoesFG}, fine-tuning has shown to give a better result in autocomplete and classification tasks. In this study, we fine-tune the \texttt{gpt-3.5-turbo} model using the training data set from Section \ref{sec:data} structuring it in the fine-tuning data format as described in the OpenAI documentation.\footnote{\url{https://platform.openai.com/docs/guides/fine-tuning}}

\paragraph{Prompt} We use customized prompt 2 in the fine-tuning setting for providing the training data. We use a second simplistic prompt after training to query the model for input data. Find the final prompt in Appendix \ref{app:prompt}.

\paragraph{Model Development} We use a similar fine-tuning parameter settings \cite{kim2023memoryefficient} to fine tuned the model.

\paragraph{Model Testing} The result of the fine-tuned model can be seen in the Table \ref{tab:finetune-table}. As we can see, performance is much lower than for the non-fine-tuned setting. However, $\alpha$ is still slightly above chance performance.  

\begin{table}[t]
    \centering
    \begin{tabular}{c|cc|cc} 
        \toprule
         \multirow{2}*{\textbf{Trial}}   & \multicolumn{2}{c|}{\textbf{dev}} & \multicolumn{2}{c}{\textbf{test}}\\     & $\alpha$ & $\%$& $\alpha$ & $\%$ \\ 
        \midrule
         1 &  0.16 & 0.24 & 0.02 & 0.12\\
         2 & 0.16  & 0.24 & 0.02 & 0.12\\
         3 & 0.16  & 0.24& 0.02& 0.12\\
         4 & 0.16  &0.24& 0.02 & 0.12\\
         5 & 0.16  &0.24& 0.02 & 0.12\\
        \midrule
        \textbf{Mean} & 0.16 & 0.24 & 0.02 & 0.12 \\
        \bottomrule
    \end{tabular}
    \vspace{0.2cm}
    \caption{Fine-tuned model results. $\alpha$: Krippendorff's $\alpha$, $\%$: Percentage agreement.}
    \label{tab:finetune-table}
\end{table}

\subsection{Automatic PhiTag prompt}

\paragraph{Prompt} By using the automatic prompting options we implemented into PhiTag, we create prompts by using the guidelines and guidelines + tutorial examples available for human annotators. We aim to see how the model performance varies by changing the prompting strategy from a manual and highly optimized to an automatic one. So, we design our prompt by concatenating an initial instruction along with the guidelines and a connecting sentence to link the tutorial examples, and a final instruction sentence. 
Additionally, we refine the guidelines by converting the data available in tables into a machine-readable format of word usages and target word. We also vary the prompt structure by changing the initial instructions message as well as the connecting sentence between guideline and tutorial examples. Further, we remove the "cannot decide" examples from the guidelines since our data does not include such instances. Similar to the customized prompt, the final instruction sentence ask the model to return a single integer value for judgment. Find the final prompts in Appendix \ref{app:prompt} and the tested prompts in the paper repository.

\paragraph{Model Development} 
We set the \texttt{top-p} and \texttt{temperature} parameters to 0.9 each, as this setting has been found to yield optimal performance in case of the customized prompt, see Section \ref{customized-prompt}. 
On five different trials, the model shows a below satisfactory performance on the development data, achieving a mean Krippendorff's $\alpha$ of -0.07 and a mean percentage agreement of 0.25 in the guidelines-only setting, and Krippendorff's $\alpha$ of 0.01 and a percentage agreement of 0.23 in the guidelines+tutorial setting, as shown in Tables \ref{tab:autoprmpt-with-guideline} and \ref{tab:autoprmpt-with-guideline-tutorial}. We notice a consistent but slight improvement of performance with modified guidelines over raw guidelines.

\begin{table}[t]
    \centering
    \begin{tabular}{c|cc|cc} 
        \toprule
         \multirow{2}*{\textbf{Trial}}   & \multicolumn{2}{c|}{\textbf{dev}} & \multicolumn{2}{c}{\textbf{test}}\\     & $\alpha$ & $\%$& $\alpha$ & $\%$ \\ 
        \midrule
         1 & -0.07 & 0.26 & 0.03 & 0.30\\
         2 & -0.07 & 0.25 & 0.03 & 0.29\\
         3 & -0.07 & 0.24 & 0.03 & 0.28\\
         4 & -0.07 & 0.25 & 0.03 & 0.28\\
         5 & -0.07 & 0.25 & 0.03 & 0.29\\
        \midrule
        \textbf{Mean} & -0.07 & 0.25 & 0.03 & 0.28 \\
        \bottomrule
    \end{tabular}
    \vspace{0.2cm}
    \caption{Automatic prompting results (guidelines-only). $\alpha$: Krippendorff's $\alpha$, $\%$: Percentage agreement.}
    \label{tab:autoprmpt-with-guideline}
\end{table}

\begin{table}[t]
    \centering
    \begin{tabular}{c|cc|cc} 
        \toprule
         \multirow{2}*{\textbf{Trial}}   & \multicolumn{2}{c|}{\textbf{dev}} & \multicolumn{2}{c}{\textbf{test}}\\     & $\alpha$ & $\%$& $\alpha$ & $\%$ \\ 
        \midrule
         1 & 0.01  & 0.23 & 0.06 & 0.33\\
         2 & 0.01 & 0.23 & 0.05& 0.31\\
         3 & 0.01  &0.23 & 0.06 &0.32 \\
         4 & 0.01  &0.23& 0.06 &0.32 \\
         5 & 0.01  &0.23& 0.06 &0.32\\
        \midrule
        \textbf{Mean} & 0.01 & 0.23 & 0.06 & 0.32 \\
        \bottomrule
    \end{tabular}
    \vspace{0.2cm}
    \caption{Automatic prompting results (guidelines and tutorials). $\alpha$: Krippendorff's $\alpha$, $\%$: Percentage agreement.}
    \label{tab:autoprmpt-with-guideline-tutorial}
\end{table}

\paragraph{Model Testing} 

The optimal settings for model-development are applied to the test data. In five successive trials the model's performance is below satisfactory, achieving a Krippendorff's $\alpha$ of 0.03 and a percentage agreement of 0.28 in the guidelines-only setting a mean Krippendorff's $\alpha$ of 0.06 and a mean percentage agreement of 0.32 in the guidelines+tutorial setting. See Table \ref{tab:autoprmpt-with-guideline} and \ref{tab:autoprmpt-with-guideline-tutorial}.

\section{Discussion}

We find that the customized prompt yields a reasonable performance while the automatic PhiTag prompting strategies show a considerably lower performance. While we were able to slightly improve the performance of the automatic prompting strategies through adjustments of the guidelines, the best model still shows low performance. A possible reason is the extensive length of the guidelines, as compared to usual prompts. In the future, it may be interesting to use only the task introduction part or the main task description part of human guidelines to prompt the model. Also, automatic summarization techniques may be used to condense the information for the model. 
Fine-tuning did not improve the performance of the customized prompt either, but rather led to a significant performance drop.

\begin{figure}[t]
\centering
\subfigure[Customized prompt]{

\includegraphics[width=0.4\textwidth]{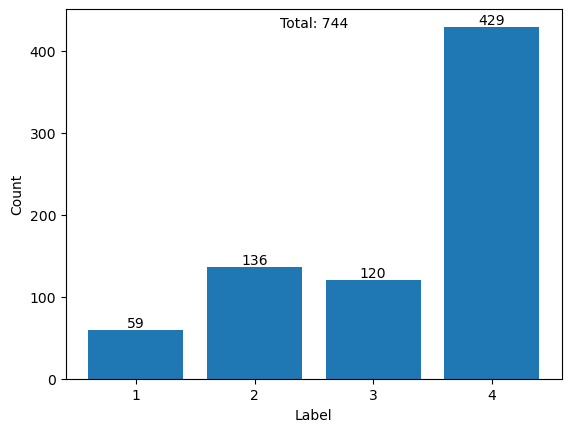}}
\qquad
\subfigure[Fine-Tuned Model]{

\includegraphics[width=0.4\textwidth]{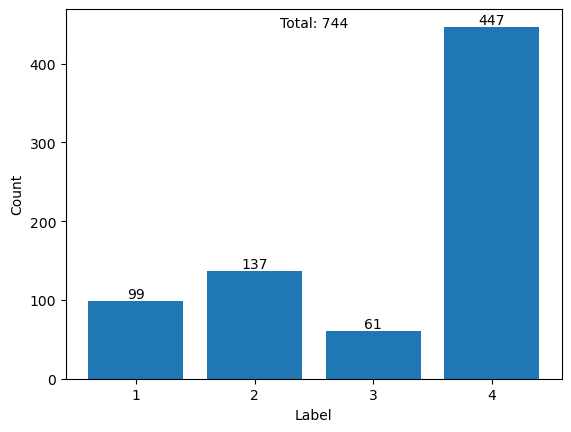}}
\subfigure[Automatic prompt (guidelines-only)]{

\includegraphics[width=0.4\textwidth]{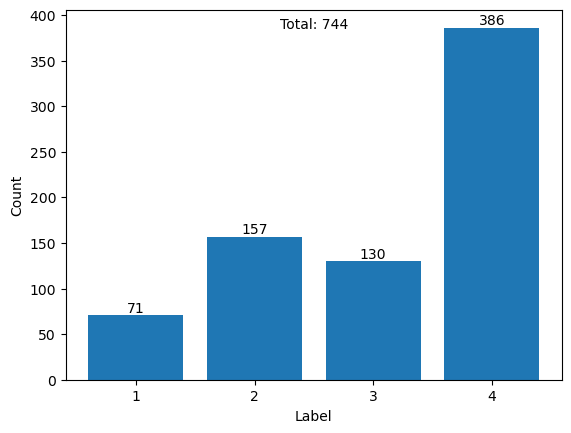}}
\qquad
\subfigure[Automatic prompt (guidelines+tutorial)]{

\includegraphics[width=0.4\textwidth]{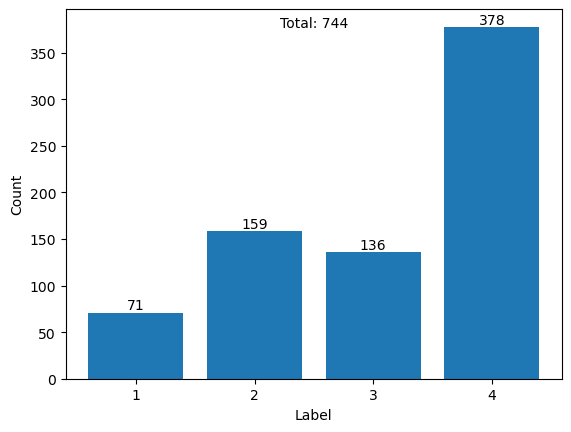}}
\caption{Distribution of predicted labels on test data.}
\label{fig:label-distribution-test}
\end{figure}

In Figure \ref{fig:label-distribution-test}, we plot the label distribution for the best model of each strategy (customized, fine-tuned, guidelines-only, guidelines+tutorial). In comparison to the distribution of the gold data (Table \ref{fig:label-distribution-data-gold}), we see that, despite partly low performance, all models recover the overall gold label distribution quite well. This holds even for the zero-shot customized prompt model having no access to the overall label distribution from the prompt in the testing phase.

We observed a significant impact of the parameters \texttt{temperature} and \texttt{top-p} on the performance of the GPT-4 model. Lower \texttt{temperatures} and led to poor performance on the development data while a \texttt{temperature} setting of 0.9 yielded optimal results.

\section{Conclusion}

In conclusion, our study reveals the crucial role of accurate prompt design and careful parameter selection in optimizing the performance of GPT for the Use Pair Semantic Proximity task. Through our experiments, we observed the significant impact of model configurations, prompting strategies, and fine-tuning techniques on enhancing the efficiency and accuracy of annotation processes. Fully automating the annotation process reusing human-tailored instructions remains a major challenge. We were not able to obtain good performance with human-like automatic prompting strategies. 

\section{Limitations}
One limitation of our study is the dependence on a specific language model and annotation platform, which may not generalize to all language models or annotation tools. Additionally, the effectiveness of our fine-tuning approach may vary depending on the task and dataset used, and further investigation with different models and datasets is needed. Furthermore, while our experiments provide insights into the effectiveness of various model configurations and prompting strategies, they may not capture the full complexity of real-world annotation scenarios.

Automatic prompts containing complete guidelines or tutorial data can get quite large and often running them costs a lot of money. Additionally, the current implementation of the automatic prompting strategy in PhiTag is rather inefficient as it annotates one instance at a time instead of creating one prompt for all instances. 

An additional limitation arising with the use of closed-source source LLMs is the question how much of the testing data the model has already seen in its training process as the data is publicly available and also used in other studies for prompting.

\bibliography{Bibliography-general,bibliography-self,additional_references}
\bibliographystyle{vancouver}

\newpage
\appendix

\section{Prompts} 
\label{app:prompt}

\paragraph{Customized Prompt 1}
 \label{ex:prompt-1}
You are a highly trained text data annotation tool capable of providing subjective responses. \\
Your task is to rate the degree of semantic relatedness between two uses of a target word in the given sentences. \\
Sentence 1: [SENTENCE 1] \\
Sentence 2: [SENTENCE 2]\\
Target word: [TARGET WORD] \\
Please provide a judgment as a single integer. For example, if your judgment is Identical, then provide 4. If your judgment is Unrelated, provide 1.

\paragraph{Customized Prompt 2}
\label{ex:prompt-2}
You are a highly trained text data annotation tool capable of providing subjective responses. \\
Rate the semantic similarity of the target word  in these sentences 1 and 2. Consider only the objects/concepts the word forms refer to: ignore any common etymology and metaphorical similarity! Ignore case! Ignore number
(cat/Cats = identical meaning). If target is emoji then rate by its contextual function. Homonyms (like bat the animal vs bat in baseball) count as unrelated. Output numeric rating: 1 is unrelated; 2 is distantly related; 3 is closely related; 4 is identical meaning.Your response should align with a human's succinct judgment. \\
Sentence 1:[SENTENCE 1]\\
Sentence 2: [SENTENCE 2]\\
Target word: [TARGET WORD] \\
Please provide a judgment as a single integer. For example, if your judgment is Identical, then provide 4. If your judgment is Unrelated, provide 1.

\paragraph{Fine-tuned model prompt}
\label{ex:prompt-5}
You are a highly trained text data annotation tool capable of providing subjective responses. \\
Annotate this pair of given sentences \\
Sentence 1: [SENTENCE 1] \\
Sentence 2: [SENTENCE 2]\\
Target word: [TARGET WORD] \\

\paragraph{Automatic PhiTag prompt with guidelines-only}
\label{ex:prompt-3}
You are a highly trained text data annotation tool capable of providing subjective responses. \\
{[}MODIFIED GUIDELINES] \\
Sentence 1: [SENTENCE 1] \\
Sentence 2: [SENTENCE 2]\\
Target word: [TARGET WORD] \\
Please provide a judgment as a single integer for Sentence 1 and Sentence 2 above. For example, if your judgment is Identical, then provide 4. If your judgment is Unrelated, provide 1.\\

\paragraph{Automatic PhiTag prompt with guidelines and tutorial}
\label{ex:prompt-4}
You are a highly trained text data annotation tool capable of providing judgments based on contexts provided to you. \linebreak
{[}MODIFIED GUIDELINES] \\
Here are few sample instances and their corresponding judgements:\\
Example sentences \\
Sentence 1: [SENTENCE 1] \\
Sentence 2: [SENTENCE 2]\\
Target word: [TARGET WORD] \\
Please provide a judgment as a single integer for Sentence 1 and Sentence 2 above. For example, if your judgment is Identical, then provide 4. If your judgment is Unrelated, provide 1.\\

\end{document}